\documentclass{ifacconf}

\usepackage[utf8]{inputenc}
\usepackage{amsmath,amssymb}
\usepackage{graphicx}
\usepackage{natbib}

\usepackage{array}        
\usepackage{booktabs}     
\usepackage{tabularx}     
\usepackage{makecell}     
\usepackage{pgfplotstable}
\pgfplotsset{compat=1.18}

\usepackage{siunitx}

\usepackage{flafter}            
\usepackage[section]{placeins}  
\usepackage{float}              
\usepackage{caption}           
\providecommand{\printnum}[1]{\pgfmathprintnumber[fixed,precision=3]{#1}}

\usepackage{needspace}

\usepackage[shortlabels]{enumitem}
\newcolumntype{Y}{>{\raggedright\arraybackslash}X}

\makeatletter
\def\@logohead{}
\makeatother

\begin{document}
\begin{frontmatter}

\title{Task-Aware Morphology Optimization of Planar Manipulators via Reinforcement Learning
	}



%
\author[DeptAffil]{Arvind Kumar Mishra}
\author[DeptAffil]{Sohom Chakrabarty}
\vspace*{-0.2cm}
\address[DeptAffil]{Department of Electrical Engineering, Indian Institute of Technology Roorkee, Roorkee 247667,India\\
	(arvind\_m@ee.iitr.ac.in; sohom@ee.iitr.ac.in)}

\begin{abstract}
\end{abstract}

\begin{abstract}
	In this work,Yoshikawa's manipulability index is adopted as the basis for investigating reinforcement learning (RL) as a framework for morphology optimization in planar robotic manipulators. A 2R manipulator tracking a circular end-effector path was first examined, since this setting admits a closed-form analytical optimum that both links are equal in length and the second joint is orthogonal to the first. This case was used to validate that the optimum could be rediscovered from the RL framework with reward feedback alone, without access to the closed-form manipulability expression or the Jacobian model. Three  RL algorithms---Soft Actor--Critic (SAC), Deep Deterministic Policy Gradient (DDPG), and Proximal Policy Optimization (PPO)---were compared against grid search and black-box optimizers, with the morphology represented by a single action parameter $\phi$ that maps to $(L_1, L_2)$ under the circular constraint. All methods converged to the analytical solution, showing that the optimum can be recovered numerically even when the analytical model is not supplied.
	
	Most morphology design problems do not admit closed-form solutions, and heuristic or grid search becomes increasingly expensive due to dimensionality and discretization effects. RL was therefore considered as a scalable alternative. The RL formulation and training setup used in the circular case were retained for elliptical and rectangular paths, while the action space were expanded from a single parameter $\phi$ to the full morphology vector $(L_1, L_2, \theta_2)$. In these non-analytical tasks, RL continued to converge reliably, whereas extending grid or black-box search to these cases would require substantially larger evaluation budgets due to the higher dimensionality of the morphology space.These results indicate that RL is suitable both for recovering known analytical optima and for morphology optimization in settings lacking closed-form solutions.
\end{abstract}

\begin{keyword}
	reinforcement learning, morphology optimization, manipulability,planar manipulator, SAC, DDPG, PPO, bandit formulation.
\end{keyword}


\end{frontmatter}

\section{Introduction}

The morphology of a manipulator—defined by its link lengths and joint configuration—plays a central role in determining its dexterity, energy efficiency, and overall task performance (Craig (2005); Siciliano et al. (2010); Yoshikawa (1985)). Yoshikawa’s manipulability index provides a classical measure of kinematic dexterity through the determinant of the Jacobian matrix (Yoshikawa (1985)). For a planar 2R manipulator, the measure of Yoshikawa gives a closed-form maximum corresponding to equal link lengths and an orthogonal joint configuration (Yoshikawa (1985); Craig (2005)). Specifically, for a circular path with the manipulator base at the centre, the analytical optimum occurs when
\[
L_1 = L_2 = \frac{R}{\sqrt{2}}, \quad \theta_2 = 90^\circ,
\]
which simultaneously maximizes manipulability and satisfies the geometric constraint \(x^2 + y^2 = R^2\) (Yoshikawa (1985)). However, for manipulators with higher degrees of freedom (DOF) or for arbitrary task trajectories, the Jacobian becomes highly nonlinear and coupled, making analytical optimization infeasible (Craig (2005)). This motivates the use of numerical and data-driven learning techniques for morphology optimization.

Traditional optimization methods such as grid search and heuristic algorithms—including particle swarm optimization (PSO) (Kennedy and Eberhart (1995)), Bayesian optimization (BO) (Mockus et al. (1978); Jones et al. (1998)), and covariance matrix adaptation evolution strategy (CMA-ES) (Hansen (2006))—have been widely employed to explore robot design spaces. Although these general-purpose algorithms can identify near-optimal configurations, they often experience discretization errors, slow convergence, and poor scalability with increasing dimensionality (Hansen (2006); Jones et al. (1998)). Furthermore, they typically lack mechanisms to handle actuator-level constraints such as joint velocity or torque limits, restricting their utility in dynamic or physically realistic scenarios (Zhang and Song (2008); De Luca and Siciliano (1992); Gros and Zanon (2020)).

Recent works have extended these heuristic techniques directly to manipulator and robot morphology optimization. For example, Farooq et al. (2021) used PSO to optimize the structural parameters of a Tricept parallel manipulator for enhanced dexterity, while Shahbazi et al. (2024) applied PSO for dynamic parameter design of Stewart manipulators. Similarly, CMA-ES has been successfully applied in soft-robot morphology design (Medvet et al. (2021)) and reviewed as a promising method in recent soft-robot optimisation surveys (Stroppa et al. (2024)). These methods, however, do not adapt well to time-varying constraints, such as changing workspace geometry, payload, or joint limits, since a full re-optimization is required whenever the conditions change.

Reinforcement Learning (RL) provides a flexible, model-free alternative capable of optimizing morphology in continuous action spaces under complex reward definitions (Sutton and Barto (2018); Haarnoja et al. (2018); Lillicrap et al. (2016); Schulman et al. (2017)). In this work, three continuous-control Deep RL algorithms—Soft Actor–Critic (SAC), Deep Deterministic Policy Gradient (DDPG), and Proximal Policy Optimization (PPO)—are utilized. The morphology variables $[L_1,\,L_2,\,\theta_2]$ are represented as continuous actions within a single-state bandit framework, allowing the agent to explore and learn the optimal geometry through reward feedback. For the circular task, RL is benchmarked against the analytical optimum (Yoshikawa (1985)) and numerical baselines to validate its applicability. Once validated, the same framework is extended to non-circular trajectories such as elliptical and rectangular paths, where analytical solutions are unavailable.

The main contributions of this work are as follows:
\begin{itemize}
	\item Verification of Yoshikawa’s manipulability optimum using RL and heuristic methods on the circular path.
	\item Demonstration that RL can serve as a reliable reference for morphology optimization when analytical solutions do not exist.
	\item Analysis of the computational and scalability limitations of grid and heuristic algorithms (Jones et al. (1998); Hansen (2006)).
	\item Illustration of how the RL framework can be extended to handle dynamic constraints and higher-DOF manipulators.
\end{itemize}

The remainder of this paper is organized as follows. Section 2 reviews related work on manipulability and morphology optimization. Section 3 presents the analytical and heuristic formulations. Section 4 describes the RL framework and reward design. Section 5 reports comparative results for circular, elliptical, and rectangular tasks. Section 6 discusses  and future scope, followed by conclusions in Section 7.

\section{Related Work}

\subsection{Manipulability and Morphology Optimization}

The concept of manipulability, introduced by Yoshikawa (1985), is still widely used in evaluating the dexterity of robotic manipulators. It relates the Jacobian determinant to the isotropy of velocity transmission in Cartesian space. Subsequent works and textbooks, including Craig (2005) and Siciliano et al. (2010), formalized the relationship between manipulator morphology, joint configuration, and task-space performance. For planar 2R manipulators, analytical optimization reveals that equal link lengths and orthogonal joint configurations maximize manipulability when the circular path is centered at the manipulator base, providing a useful benchmark for validating computational methods.

\subsection{Heuristic and Evolutionary Optimization}

To overcome the limitations of analytical formulations, many studies have adopted numerical and heuristic optimization methods. Classical global optimizers such as particle swarm optimization (PSO) (Kennedy and Eberhart (1995)), Bayesian optimization (BO) (Mockus et al. (1978); Jones et al. (1998)), and covariance matrix adaptation evolution strategy (CMA-ES) (Hansen (2006)) have been widely applied to robot design and parameter tuning. Although these general-purpose techniques can identify near-optimal configurations, they typically incur high computational cost, slow convergence, and poor scalability with increasing design dimensionality (Hansen (2006); Jones et al. (1998)). Furthermore, they often neglect actuator-level constraints such as joint velocity and torque limits, limiting their applicability to dynamic systems (Zhang and Song (2008); De Luca and Siciliano (1992); Gros and Zanon (2020)). Recent works have extended these methods to manipulator morphology, such as PSO-based optimization of parallel and Stewart platforms (Shahbazi et al. (2024)) and CMA-ES-driven design of soft robotic structures (\citep{MedvetVoxelSoft2021,Stroppa2024}). While these studies demonstrate that heuristic algorithms can adapt to manipulator design, their reliance on dense sampling makes them computationally expensive and unsuitable for real-time or adaptive morphology optimization.The specific configurations and hyperparameters used for PSO, BO, and CMA-ES in this study are summarized in Table~\ref{tab:heuristic-params}.

\vspace*{-0.05cm}

\subsection{Learning-Based and Reinforcement Learning\\Approaches}

With the rise of deep reinforcement learning (DRL), several studies have explored learning-based co-design, where morphology and control are optimized jointly. RL provides a model-free formulation capable of handling continuous actions and complex rewards (Sutton and Barto (2018)). Algorithms such as SAC (Haarnoja et al. (2018)), DDPG (Lillicrap et al. (2016)), and PPO (Schulman et al. (2017)) enable continuous optimization without explicit gradient derivations of the kinematic or dynamic models. Learning-based co-design frameworks have been applied in areas such as tool design, legged locomotion, and soft-robot morphology evolution; these studies show reinforcement learning can scale effectively to complex design problems and adapt better to changing conditions than traditional heuristic methods (Haarnoja et al. (2018); Lillicrap et al. (2016); Schulman et al. (2017)).

\subsection{Motivation and Scope}

Despite these advances, a unified comparison between analytical, heuristic, and reinforcement learning approaches for morphology optimization remains limited. In particular, the classical Yoshikawa benchmark has rarely been revisited under a learning-based framework. This study bridges that gap by verifying Yoshikawa’s analytical result using both numerical and RL methods on a circular task, then extending the framework to non-circular trajectories where closed-form solutions are unavailable.

\section{Analytical and Heuristic Formulation}

\subsection{Analytical Benchmark: Planar 2R Manipulator}

\begin{figure}[t]
	\centering
	\includegraphics[width=0.55\linewidth]{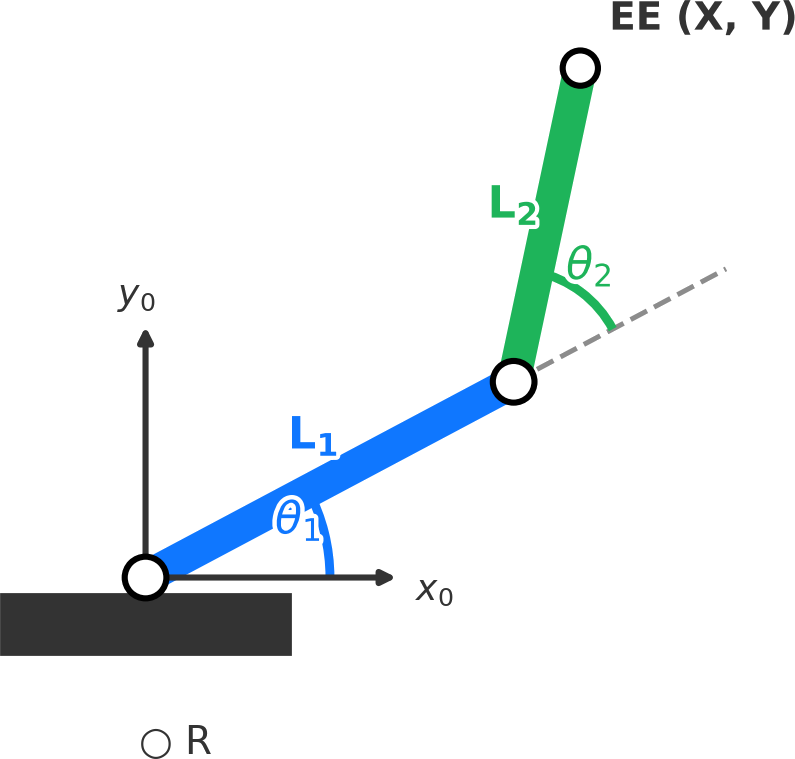}
	\caption{Planar 2R manipulator schematic.}
	\label{fig:2R-schematic}
\end{figure}

The planar two–revolute (2R) manipulator in Fig.~\ref{fig:2R-schematic} is used as the analytical benchmark for morphology optimization.  
Its end–effector Cartesian coordinates are given by the standard forward kinematics
\begin{subequations}\label{eq:fwd-kin}
	\begin{align}
		x &= L_1 \cos\theta_1 + L_2 \cos(\theta_1+\theta_2), \\
		y &= L_1 \sin\theta_1 + L_2 \sin(\theta_1+\theta_2),
	\end{align}
\end{subequations}
where $L_1,L_2$ are the link lengths and $\theta_1,\theta_2$ are the joint angles.

The corresponding Jacobian is
\begin{equation}\label{eq:jacobian}
	J(\theta)=
	\begin{bmatrix}
		-L_1\sin\theta_1 - L_2\sin(\theta_1+\theta_2) & -L_2\sin(\theta_1+\theta_2)\\[3pt]
		L_1\cos\theta_1 + L_2\cos(\theta_1+\theta_2) &  L_2\cos(\theta_1+\theta_2)
	\end{bmatrix}.
\end{equation}

%
%

Yoshikawa’s manipulability index (Yoshikawa (1985)) measures the isotropy of velocity transmission. 
For a planar 2R manipulator, the determinant of the Jacobian simplifies to
\begin{equation}\label{eq:manip}
	w(\theta_2)=\sqrt{\det(JJ^{\!\top})}=L_1L_2\lvert\sin\theta_2\rvert .
\end{equation}
\vspace*{-0.9cm}
\paragraph{Closed-form optimum for a centered circular task.}
Consider a circular end-effector trajectory of radius $R$ centred at the base. 
Maximum manipulability is achieved when $\lvert\sin\theta_2\rvert$ attains its maximum value,
\[
\theta_2 = \ang{90} \quad \Rightarrow \quad \lvert\sin\theta_2\rvert = 1 .
\]
Under this configuration, the reachability constraint reduces to
\begin{equation}\label{eq:reach}
	L_1^2 + L_2^2 = R^2 ,
\end{equation}
which follows from the geometry of the manipulator at full extension.

Maximizing~\eqref{eq:manip} subject to~\eqref{eq:reach} yields the symmetric solution
\[
L_1 = L_2 = \frac{R}{\sqrt{2}},
\]
since the product $L_1L_2$ is maximized when both links are equal. 
Substituting this result into~\eqref{eq:manip} gives the peak manipulability
\begin{equation}\label{eq:wmax}
	w_{\max} = R^2/2.
\end{equation}

This equal-length configuration constitutes the analytical optimum for the centered circular path and is used as a \emph{baseline morphology} throughout our experiments. 
For non-circular, off-centre, or constrained workspaces, no closed-form optimum exists, motivating the numerical and reinforcement-learning methods developed in the remainder of the paper.

\subsection{Grid-Search Verification}

To validate the analytical optimum derived in Section~3.1, a one-dimensional sweep over the morphology parameter $\phi$ is performed instead of a full grid search over $(L_1,L_2,\theta_2)$. For a circular end-effector path of fixed radius $R$, the variables are related by the inverse-kinematics (IK) constraint
\begin{equation}\label{eq:circle-constraint}
	R^2 = L_1^2 + L_2^2 + 2 L_1 L_2 \cos\theta_2 .
\end{equation}
Substituting~\eqref{eq:circle-constraint} into the manipulability expression~\eqref{eq:manip} yields a form that depends only on the link lengths:
\begin{equation}\label{eq:w-l1l2}
	w(L_1,L_2) = L_1L_2 \sqrt{1 - 
		\left( \frac{R^2 - L_1^2 - L_2^2}{2L_1L_2} \right)^2 } .
\end{equation}
This expression is included for completeness, as it corresponds to the full two-dimensional grid search over $(L_1,L_2)$ before the problem is reduced using the circular reachability constraint.
\vspace*{-0.6cm}
%

\begin{figure}[!htb]
	\centering
	\includegraphics[width=0.45\linewidth]{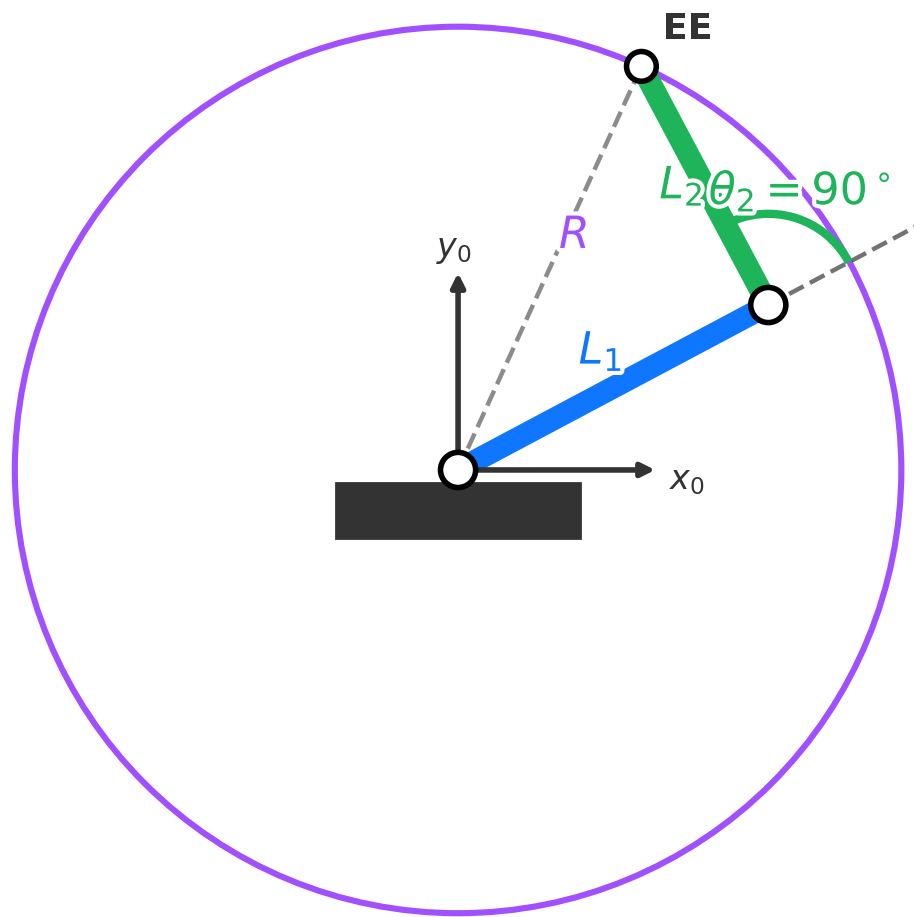}
	\caption{Maximum-manipulability configuration of the planar 2R manipulator for a centered circular end-effector path ($L_1 = L_2 = R/\sqrt{2}$, $\theta_2 = 90^\circ$).}
	\label{fig:2Rcircle}
\end{figure}

\paragraph{Reduced one-dimensional sweep.}
Since the reachability condition given by Eq.~\eqref{eq:reach} restricts feasible designs to lie on a circular locus in the $(L_1, L_2)$ plane (for $L_1, L_2 > 0$), the morphology can be parameterized by a single variable:
\begin{equation}\label{eq:l1l2-phi}
	L_1 = R\cos\phi, 
	\qquad 
	L_2 = R\sin\phi,
	\qquad 
	\phi \in [0,\tfrac{\pi}{2}] .
\end{equation}
Substituting~\eqref{eq:l1l2-phi} into~\eqref{eq:manip} gives the normalized manipulability
\begin{equation}\label{eq:wnorm-phi}
	\bar{w}_{\mathrm{norm}} = \lvert \sin(2\phi) \rvert .
\end{equation}
A sweep over $\phi$ confirms that the maximum occurs at
\[
\phi = 45^{\circ}
\quad\Rightarrow\quad
L_1 = L_2 = R/\sqrt{2},
\]
which matches the link-length solution obtained analytically. 
The corresponding joint angle $\theta_2 = 90^{\circ}$ follows from the analytical maximization in Section~3.1 and is shown here only for completeness, since it is not recovered from the one-dimensional sweep itself. 
The resulting configuration is visualized in Fig.~\ref{fig:2Rcircle}.

\FloatBarrier

\subsection{Heuristic Baselines}

Before introducing the reinforcement learning approach, three commonly used black-box optimization methods are considered for comparison: Particle Swarm Optimization (PSO) (Kennedy and Eberhart (1995)), Bayesian Optimization (BO) (Mockus (1978); Jones et al. (1998)), and the Covariance Matrix Adaptation Evolution Strategy (CMA-ES) (Hansen and Ostermeier (2006)).  
Each method searches over the link lengths $(L_1, L_2)$ in order to maximize the average normalized manipulability along the specified end-effector path.

PSO is a population-based method in which a swarm of particles explores the design space and updates candidate solutions by combining personal best and global best information.  
BO constructs a probabilistic surrogate of the objective function and selects new samples through an acquisition rule that trades off exploration and exploitation.  
CMA-ES maintains a multivariate Gaussian distribution over the design variables and adapts its mean and covariance iteratively, enabling efficient sampling in regions of high performance.

These methods work well for the present 2-D circular morphology problem but require many objective evaluations and repeated inverse-kinematics calls.  
Their computational cost grows rapidly when additional morphology parameters are introduced (e.g., link offsets, joint limits, payload effects) or when the task is uncertain or time-varying (Zhang et al. (2008); De Luca and Oriolo (1992); Gros and Ferreau (2020)).  

For this reason, RL is considered as a scalable alternative: once trained, a policy can output morphology parameters directly, without re-running an optimizer for every new task instance.  
In our experiments, PSO, BO, and CMA-ES are evaluated only on the circular case, where the analytical optimum is known and provides a ground-truth reference.  
The hyperparameters used for these heuristic baselines are listed in Table~\ref{tab:heuristic-params}.

\section{Reinforcement Learning Framework}
\label{sec:rl-framework}

\subsection{Problem Formulation}
\label{sec:rl-formulation}
The morphology–optimization task is formulated as a single-step reinforcement learning (RL) problem. 
Since no temporal evolution is involved, the setting reduces to a \emph{contextual bandit}, in which a geometry is selected and an immediate reward is returned.
The state $s$ represents the task path (e.g., circular, elliptical, or rectangular), and the action corresponds to the manipulator geometry,
\begin{equation}
	a \;=\; [L_1,\,L_2,\,\theta_2]^{\!\top} \;\in\; \mathcal{A} \subset \mathbb{R}^3 .
	\label{eq:action}
\end{equation}

It is noted that this full 3-D action space is required only for non-circular tasks, where no closed-form optimum exists and manipulability depends jointly on $(L_1,L_2,\theta_2)$.  
For the circular task, the action dimensionality is reduced to a single parameter $\phi$, since
\[
L_1 = R\cos\phi,\qquad L_2 = R\sin\phi,\qquad \theta_2 = 90^\circ
\]
are already known from the analytical solution.  
In that case, the RL agent selects only $\phi$, and the task becomes a one-dimensional bandit.

The objective of learning is expressed as
\begin{equation}
	J(\pi)\;=\;\mathbb{E}_{s\sim\mathcal{D},\,a\sim\pi(\cdot\,|\,s)}\!\big[r(s,a)\big],
	\label{eq:objective}
\end{equation}

where $\pi(\cdot|s)$ denotes the policy that outputs the geometry for a given task context (Sutton and Barto (2018)).

\noindent\textbf{Notation for \eqref{eq:objective}.}
\begin{itemize}
	\item $s\in\mathcal{S}$: task context (path descriptor). In this work it encodes the path type and parameters
	(e.g., circle radius $R$, ellipse $(a_e,b_e)$, or rectangle $(w,h)$).
	\item $a\in\mathcal{A}$: action (geometry). For non-circular tasks, $a=[L_1,L_2,\theta_2]^{\!\top}$.
	For the circular task, the action reduces to a single parameter $a=[\phi]$ with
	$L_1=R\cos\phi$, $L_2=R\sin\phi$, and $\theta_2=90^\circ$.
	\item $\mathcal{D}$: distribution over task contexts used during training (in our experiments, one
	context per experiment; see Section~\ref{sec:reward-design} for how the path is sampled to
	compute rewards).
	\item $\pi(\cdot|s)$: (stochastic) policy over actions given context $s$, parameterized by neural
	networks (SAC/DDPG/PPO) and implemented with $\tanh$-squashed outputs mapped to physical ranges.
	For DDPG the executed action is deterministic with added exploration noise during training.
	\item $r(s,a)$: scalar reward defined in Section~\ref{sec:reward-design}
	(one of $r_{\mathrm{raw}}$, $r_{\mathrm{norm}}$, $r_{\mathrm{band}}$, or $r_{\mathrm{hyb}}$).
\end{itemize}


\noindent The expectation in \eqref{eq:objective} is taken over $s \sim \mathcal{D}$ and
$a \sim \pi(\cdot|s)$ (policy stochasticity and, when applicable, exploration noise) (Sutton and Barto (2018)).


\subsection{Reward Design}
\label{sec:reward-design}

\begin{figure}
	\centering
	
	{
		\includegraphics[width=0.45\linewidth]{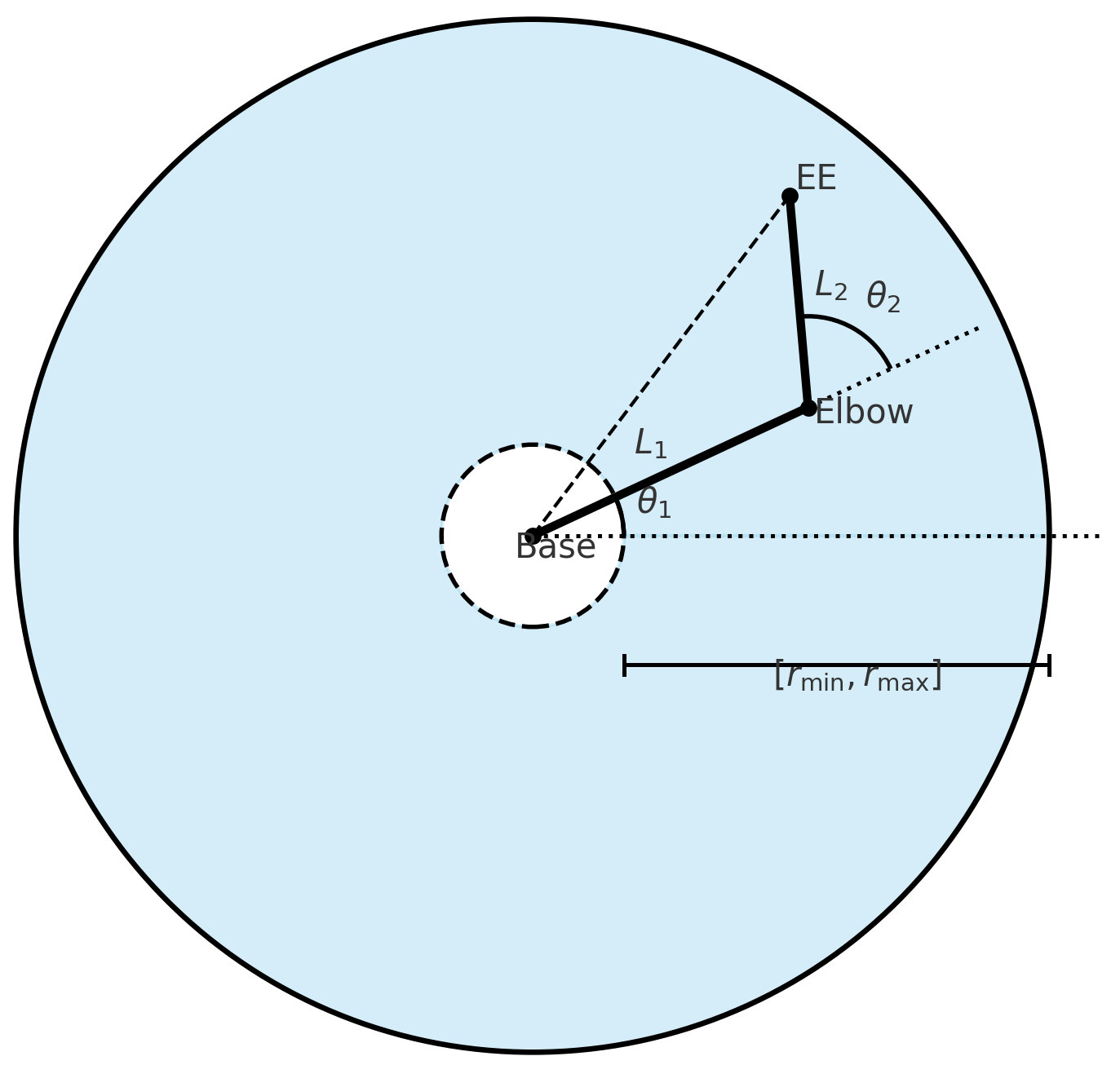}%
	}
	\caption{Reachable annulus of the planar 2R manipulator showing link lengths
		$L_1$, $L_2$, joint angles $\theta_1$, $\theta_2$ (elbow-up), and the
		inner/outer radii $[r_{\min}, r_{\max}]$. The shaded band corresponds to the geometric annulus that is reachable in principle for the chosen $(L_1,L_2)$.}
	\label{fig:annulus}
\end{figure}

The reward is computed by evaluating reachability and manipulability over $N$ uniformly sampled points on the task path,
$P=\{(x_k,y_k)\}_{k=1}^N$.
Let $r_k=\sqrt{x_k^2+y_k^2}$ denote the radial distance of the $k$-th point from the base, and define the morphology radii
\[
r_{\min}=|L_1-L_2|,\qquad r_{\max}=L_1+L_2.
\]
A point is considered reachable if it lies within this annulus and the elbow-up inverse kinematics returns a valid joint configuration within limits, as illustrated in Fig.~\ref{fig:annulus}:
\[
\mathbb{I}_k=
\begin{cases}
	1, & \parbox[t]{.62\columnwidth}{%
		$r_k\in[r_{\min},r_{\max}]$ and elbow-up IK returns $(\theta_{1,k},\theta_{2,k})$ within limits%
	}\\[4pt]
	0, & \text{otherwise.}
\end{cases}
\]

The number of reachable samples and the corresponding coverage are
\[
N_{\text{reach}}=\sum_{k=1}^N \mathbb{I}_k,\qquad
\mathrm{cov}=\frac{N_{\text{reach}}}{N}.
\]

For all reachable targets ($\mathbb{I}_k{=}1$), the absolute and normalized manipulability values are
\[
w_k=\big|L_1L_2\sin\theta_{2,k}\big|,\qquad
w^{\mathrm n}_k=\big|\sin\theta_{2,k}\big|,
\]
and their averages are
\[
\bar w=\frac{1}{N_{\text{reach}}}\sum_{k=1}^N \mathbb{I}_k\,w_k,\qquad
\bar w_{\mathrm n}=\frac{1}{N_{\text{reach}}}\sum_{k=1}^N \mathbb{I}_k\,w^{\mathrm n}_k,
\]
with both set to zero if $N_{\text{reach}}=0$.

\smallskip
\noindent\textbf{Meaning of terms used above.}
\begin{itemize}
	\item \textbf{\boldmath$N$}: number of task samples; $P$ is the set of sampled Cartesian targets.
	\item \textbf{\boldmath$r_k$}: radial distance of sample $k$; used to check geometric reachability.
	\item \textbf{\boldmath$r_{\min},r_{\max}$}: inner/outer radii of the manipulator annulus induced by $(L_1,L_2)$.
	\item \textbf{\boldmath$\mathbb{I}_k$}: binary reachability indicator for sample $k$ (1 if reachable, 0 otherwise).
	\item \textbf{\boldmath$N_{\text{reach}}$}: number of reachable targets, i.e.\ $\sum_k \mathbb{I}_k$.
	\item \textbf{\boldmath$\mathrm{cov}$}: coverage ratio $N_{\text{reach}}/N$, measuring what fraction of the task is reachable.
	\item \textbf{\boldmath$\theta_{2,k}$}: elbow angle from IK for sample $k$ (elbow-up convention).
	\item \textbf{\boldmath$w_k$}: absolute manipulability at sample $k$, incorporating link-scale and configuration.
	\item \textbf{\boldmath$w^{\mathrm n}_k$}: normalized manipulability (configuration only, scale removed).
	\item \textbf{\boldmath$\bar w$, $\bar w_{\mathrm n}$}: mean values of $w_k$ and $w^{\mathrm n}_k$ over all reachable targets.
\end{itemize}

\medskip

\noindent\textbf{Band constraints.}
Each task is associated with a desired radial band $[b,a]$:
\begin{itemize}
	\item circle (radius $R$): $[b,a]=[R,R]$,
	\item ellipse (semi-axes $a_e,b_e$):\\ $[b,a]=[\min(a_e,b_e),\,\max(a_e,b_e)]$,
	\item rectangle (width $w$, height $h$):\\ $[b,a]=\big[\tfrac12\min(w,h),\,\tfrac12\sqrt{w^2+h^2}\big]$.
\end{itemize}
Band violation is penalized by
\[
B \;=\; W_{\mathrm{in}}\,[\,b-r_{\min}\,]_+^{2} \;+\; W_{\mathrm{out}}\,[\,r_{\max}-a\,]_+^{2},
\]
where $[x]_+=\max(x,0)$ ensures that penalties apply only when the annulus is too small or too large.
Additional penalties are defined as
\[
\Delta_{\text{unr}}=W_{\mathrm{unr}}(1-\mathrm{cov}),\qquad
\Delta_{\text{len}}=W_{\mathrm{len}}(L_1+L_2).
\]
(Default weights are given in Table~\ref{tab:rl-hyperparams}.)

\medskip
\noindent\textbf{Reward functions.}
\begingroup
\setlength{\abovedisplayskip}{3pt}\setlength{\belowdisplayskip}{3pt}

\begin{align*}
	r_{\mathrm{raw}}  &=\; \bar w_{\phantom{n}} \;-\; \Delta_{\text{unr}} \;-\; \Delta_{\text{len}}, \\
	r_{\mathrm{norm}} &=\; \bar w_{\mathrm{n}} \;-\; \Delta_{\text{unr}} \;-\; \Delta_{\text{len}}, \\
	r_{\mathrm{band}} &=\; -\,B \;\;\;\;-\; \Delta_{\text{unr}} \;-\; \Delta_{\text{len}}, \\
	r_{\mathrm{hyb}}  &=\; \bar w_{\mathrm{n}} \;-\; B \;-\; \Delta_{\text{unr}} \;-\; \Delta_{\text{len}}.
\end{align*}

\endgroup

\noindent\textbf{Interpretation of penalties and rewards.}
\begin{itemize}
	\item \textbf{\boldmath$\Delta_{\text{unr}}$ (Coverage penalty)}: proportional to the fraction of unreachable points. This term discourages morphologies that leave large portions of the task unreachable; its weight $W_{\mathrm{unr}}$ sets how strongly infeasible designs are penalized.
	
	\item \textbf{\boldmath$\Delta_{\text{len}}$ (Length regularizer)}: penalizes large total link length $(L_1+L_2)$. This avoids trivial solutions that artificially increase manipulability by enlarging both links; $W_{\mathrm{len}}$ controls the regularization strength.
	
	\item \textbf{\boldmath$B$ (Band penalty)}: enforces that the morphology annulus $[r_{\min},r_{\max}]$ matches the task radial band $[b,a]$. The inner term $W_{\mathrm{in}}[b-r_{\min}]_+^2$ penalizes annuli that are too small, while the outer term $W_{\mathrm{out}}[r_{\max}-a]_+^2$ penalizes those that are too large.
	
	\item \textbf{\boldmath$r_{\mathrm{raw}}$}: rewards absolute manipulability $\bar w$ while penalizing unreachability and overly long links. It is useful when absolute dexterity (scale included) matters, e.g., the centered circular task.
	
	\item \textbf{\boldmath$r_{\mathrm{norm}}$}: rewards normalized manipulability $\bar w_{\mathrm n}$ to emphasize configuration-dependent dexterity while removing scale effects; useful when link scale should not be preferred.
	
	\item \textbf{\boldmath$r_{\mathrm{band}}$}: ignores manipulability and prioritizes geometric feasibility (coverage and band matching); useful when strict adherence to the task band is required.
	
	\item \textbf{\boldmath$r_{\mathrm{hyb}}$}: combines normalized manipulability with coverage and band penalties, providing a balanced objective for non-circular tasks. This hybrid reward is used in the experiments where manipulability must be traded off against reachability and geometry.
\end{itemize}

The weight hyperparameters and algorithm-specific settings are given in Table~\ref{tab:rl-hyperparams}.

%
%
%

\subsection{Learning Algorithms}
\label{sec:learning-algos}

Soft Actor–Critic (SAC) (Haarnoja et al. (2018)), Deep Deterministic Policy Gradient (DDPG) (Lillicrap et al. (2016)), and Proximal Policy Optimization (PPO) (Schulman et al. (2017)) are employed as representative state-of-the-art continuous-control algorithms.  
SAC and DDPG are trained off-policy using replay buffers and target critics, whereas PPO is trained on-policy using clipped surrogate objectives.  
All policies are implemented as multilayer perceptrons (MLPs) with $\tanh$ outputs, which are linearly mapped to the physical geometry ranges
$L_i \in (L_{\min},L_{\max})$ and $\theta_2 \in (0,\pi)$.\footnote{Actions in $(-1,1)$ are squashed by $\tanh$ and then scaled to their physical limits in the implementation.}

For the circular task, the analytical optimum is known and the action can be reduced to the single parameter $\phi$ (i.e., $L_1 = R\cos\phi$, $L_2 = R\sin\phi$, $\theta_2 = 90^\circ$).  
All three agents reliably recover the closed-form optimum, confirming the correctness of the RL setup.

For the elliptical and rectangular tasks, no closed-form solution exists and the full 3-D action space $(L_1,L_2,\theta_2)$ is used.  
In these cases, the hybrid reward $r_{\mathrm{hyb}}$ is adopted to balance manipulability with band feasibility, and the policies learn morphologies that maximize dexterity while satisfying geometric constraints.  
The algorithm-specific hyperparameters and reward weights are summarized in Table~\ref{tab:rl-hyperparams}.

\begin{table}[!htbp]
	\centering
	\caption{Hyperparameters used for reinforcement-learning algorithms (SAC, DDPG, PPO). Reward weights are shared across algorithms.}
	\label{tab:rl-hyperparams}
	\small
	\setlength{\tabcolsep}{5pt}
	\begin{tabular}{lccc}
		\toprule
		\textbf{Parameter} & \textbf{SAC} & \textbf{DDPG} & \textbf{PPO} \\
		\midrule
		Episodes & 5000 & 5000 & 5000 \\
		Batch size & 256 & 256 & 128 \\
		Learning rate (actor/critic) & $3\times10^{-4}$ & $1\times10^{-3}$ & $3\times10^{-4}$ \\
		Discount factor $\gamma$ & 0.99 & 0.99 & 0.99 \\
		Replay buffer size & $10^5$ & $10^5$ & --- \\
		Policy update ratio (PPO) & --- & --- & 0.2 \\
		Target smoothing $\tau$ & 0.005 & 0.005 & --- \\
		Entropy coefficient $\alpha$ (SAC) & 0.2 & --- & --- \\
		Optimizer & Adam & Adam & Adam \\
		\addlinespace
		\multicolumn{4}{l}{\textit{Reward weights (shared):}} \\
		$W_{\mathrm{unr}}$ (unreachability) & \multicolumn{3}{c}{5.0} \\
		$W_{\mathrm{in}}$ (inner-band)      & \multicolumn{3}{c}{5.0} \\
		$W_{\mathrm{out}}$ (outer-band)     & \multicolumn{3}{c}{5.0} \\
		$W_{\mathrm{len}}$ (length regularizer) & \multicolumn{3}{c}{0.5} \\
		\bottomrule
	\end{tabular}
\end{table}

\FloatBarrier

\section{Results}
\label{sec:results}

\subsection{Experimental setup: circle (analytical case)}
\label{sec:circle-setup}

For a centered circle of radius $R$, we use a scalar action $u\in(-1,1)$ mapped to
$\phi\in(\varepsilon,\tfrac{\pi}{2}-\varepsilon)$ via
$\phi(u)=\tfrac{\pi}{4}+(\tfrac{\pi}{4}-\varepsilon)u$, with $\varepsilon=10^{-3}$\,rad.
The link lengths are $L_1=R\cos\phi$, $L_2=R\sin\phi$, yielding the analytical normalized reward
$w_{\text{norm}}(\phi)=\sin(2\phi)$ (no band penalty). The analytic maximizer is
$\phi^\star=45^\circ$ with $L_1=L_2=R/\sqrt{2}$ and $w_{\text{norm}}^{\max}=1$.

\begin{table}[H]  
		\centering
		\caption{Settings for the circle benchmark (one degree of freedom).}
		\label{tab:circle-settings}
		\setlength{\tabcolsep}{3pt}\footnotesize
		\begin{tabularx}{\columnwidth}{l >{\raggedright\arraybackslash}X}
			\toprule
			\textbf{Item} & \textbf{Value} \\
			\midrule
			Path geometry & Circle of radius $R=0.40$\,m (centered at base) \\
			Action/param  & Scalar $u\in(-1,1)$, mapped to $\phi(u)$ with $\varepsilon=10^{-3}$ \\
			Lengths       & $L_1=R\cos\phi$, $L_2=R\sin\phi$ \\
			Reward        & $w_{\mathrm{norm}}=\sin(2\phi)$ (no band penalty) \\
			Analytic ref. & $\phi^\star=45^\circ$, $L_1=L_2=\tfrac{R}{\sqrt{2}}$, $w_{\mathrm{norm}}^{\max}=1$ \\
			Training      & 5 seeds; 5000 episodes; batch 256 \\
			\bottomrule
		\end{tabularx}
	\end{table}

\begin{figure}[H]
	\centering
	\includegraphics[width=\linewidth]{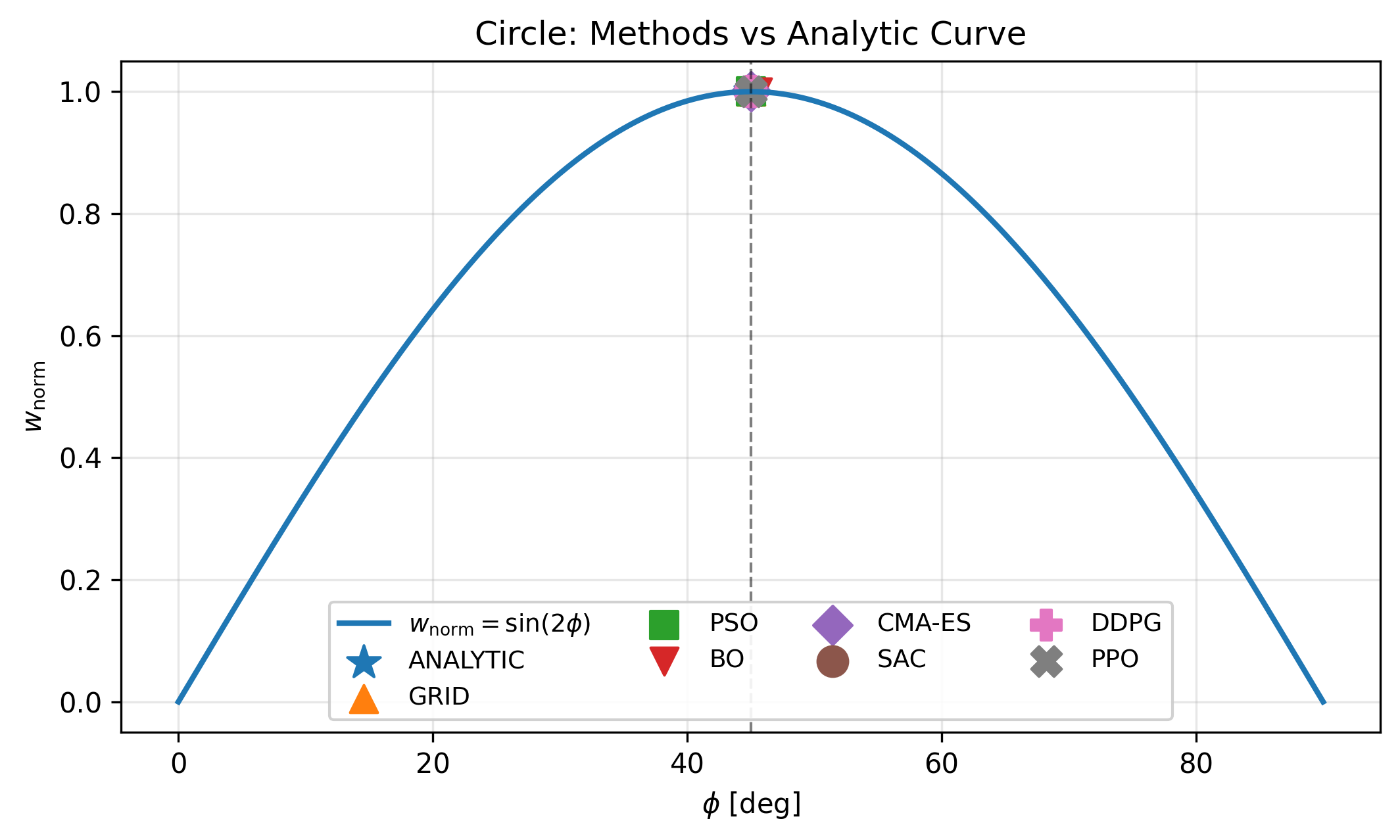}
	\caption{Circle (analytical reward): method endpoints vs.\ analytical curve
		$w_{\text{norm}}=\sin(2\phi)$. Dashed: $\phi=45^\circ$.}
	\label{fig:circle-curve}
\end{figure}

\begin{figure}
	\centering
	
	\includegraphics[width=\linewidth]{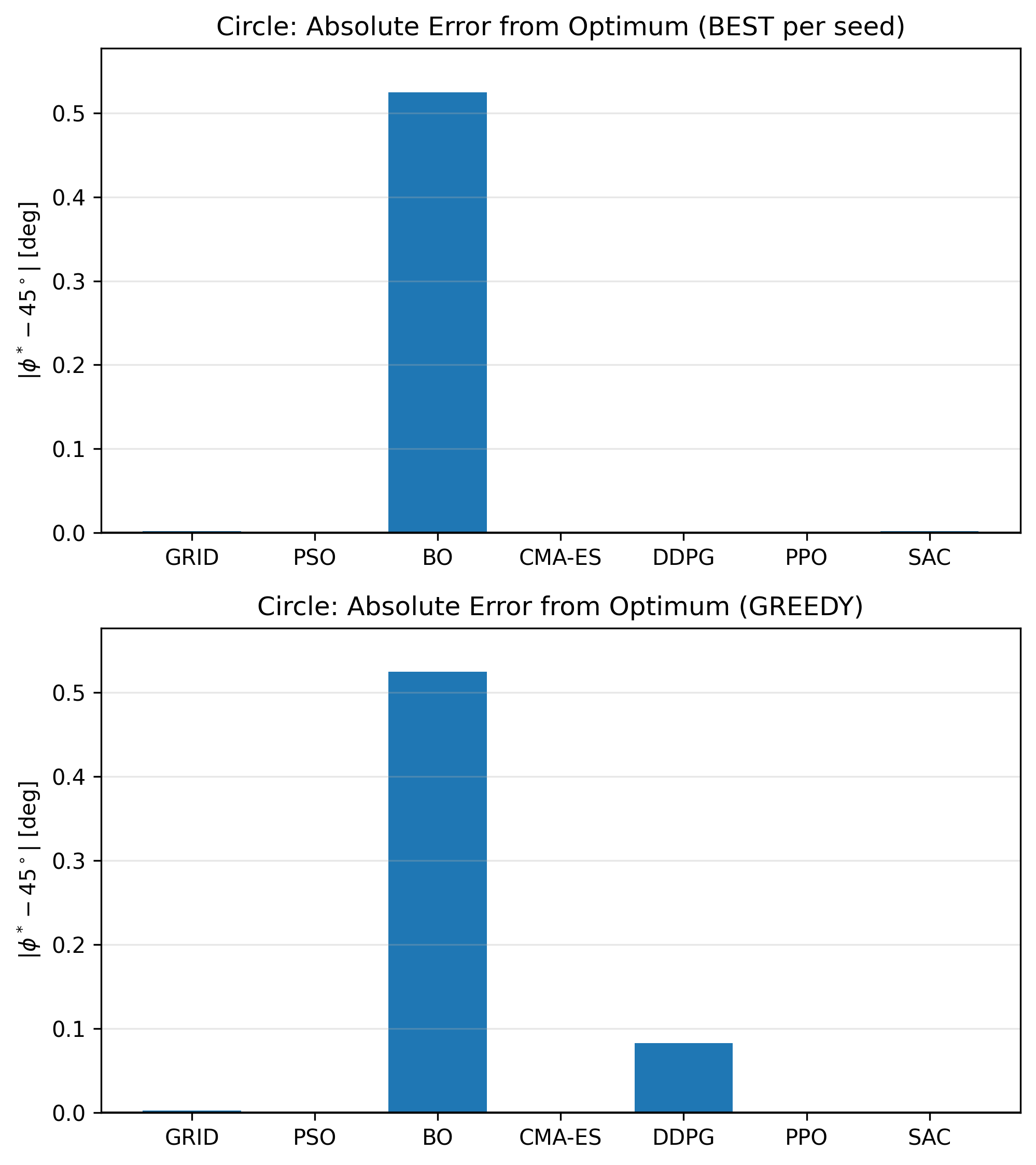}
	\caption{Circle (analytical reward): absolute deviation $|\phi^\star-45^\circ|$ (deg).
		Top: BEST; Bottom: GREEDY.}
	\label{fig:circle-abs}
\end{figure}

\pgfplotstableread[col sep=comma,trim cells=true]{circle_summary_methods.csv}\circsum
\pgfplotstablecreatecol[
create col/expr={abs(\thisrow{phi_deg}-45)}
]{abs_err_deg}\circsum

\begin{table}[t]
	\caption{Circle benchmark: summary across methods}
	\pgfplotstabletypeset[
	font=\footnotesize,
	every head row/.style={before row=\toprule, after row=\midrule},
	every last row/.style={after row=\bottomrule},
	columns={Method,phi_deg,L1_m,L2_m,w_norm},
	columns/Method/.style={string type, column name=Method, column type=l},
	columns/phi_deg/.style={column name={$\phi^\star$ [deg]}, fixed, precision=3},
	columns/L1_m/.style={column name={$L_1$ [m]}, fixed, precision=5},
	columns/L2_m/.style={column name={$L_2$ [m]}, fixed, precision=5},
	columns/w_norm/.style={column name={$w_{\text{norm}}$}, fixed, precision=6},
	column type=r,
	]{\circsum}

	\label{tab:circle-summary}
\end{table}

\FloatBarrier
\vspace*{6pt} 
\subsection{Experimental setup: ellipse/rectangle}

\begin{table}[H] 
	\centering
	\caption{Settings used for ellipse/rectangle.}
	\label{tab:res-settings}
	\setlength{\tabcolsep}{4pt}\footnotesize
	\begin{tabularx}{\columnwidth}{l >{\raggedright\arraybackslash}X}
		\toprule
		\textbf{Item} & \textbf{Value} \\
		\midrule
		Path geometry & Ellipse: $a_e=\mathbf{0.40}$\,m, $b_e=\mathbf{0.25}$\,m \quad/\quad
		Rectangle: $w=\mathbf{0.70}$\,m, $h=\mathbf{0.40}$\,m \\
		Sampling/IK   & $N=\mathbf{720}$ points, elbow-up \\
		Action bounds & $L_i\in[\mathbf{0.05},\,\mathbf{0.60}]$\,m (via $\tanh$ box) \\
		Band targets  & Ellipse: $[b,a]=[\min(a_e,b_e),\,\max(a_e,b_e)]$;\; Rectangle:
		$[b,a]=[\tfrac12\min(w,h),\,\tfrac12\sqrt{w^2+h^2}]$ \\
		\addlinespace[1pt]
		Baselines     & \textit{Equal-dex:} $L_1=L_2=L^\star$, $L^\star=\sqrt{\mathbb{E}[r^2]/2}$, with
		$\mathbb{E}[r^2]_{\mathrm{ellipse}}=\tfrac12(a_e^2+b_e^2)$ and
		$\mathbb{E}[r^2]_{\mathrm{rect}}=\dfrac{w^3+3wh^2+3hw^2+h^3}{12(w+h)}$
		(derivations in App.~\ref{app:Er2}).\\
		& \textit{Band-match:} for band $[b,a]$, take
		$L_1=\tfrac{a+b}{2}$,\; $L_2=\tfrac{a-b}{2}$ (either order). \\
		Seeds/Training & Seeds $=\{1,2,3,4,5\}$; episodes $=\mathbf{5000}$; batch $=\mathbf{256}$ \\
		\bottomrule
	\end{tabularx}
\end{table}


Across ellipse and rectangle, we use a shared protocol and the hybrid reward (Sec.~\ref{sec:reward-design}).
Band targets follow geometry; actions are box–constrained to $L_i\!\in[0.05,0.60]$\,m via $\tanh$; 
$N\!=\!720$ path points are sampled with elbow–up IK.

\subsubsection{Ellipse: results}
\label{sec:ellipse-results}


\begingroup
\setlength{\abovecaptionskip}{2pt}
\setlength{\belowcaptionskip}{2pt}
\renewcommand{\arraystretch}{0.96}
\setlength{\tabcolsep}{3pt}

\begin{center}
	\includegraphics[width=\columnwidth,height=0.28\textheight,keepaspectratio]{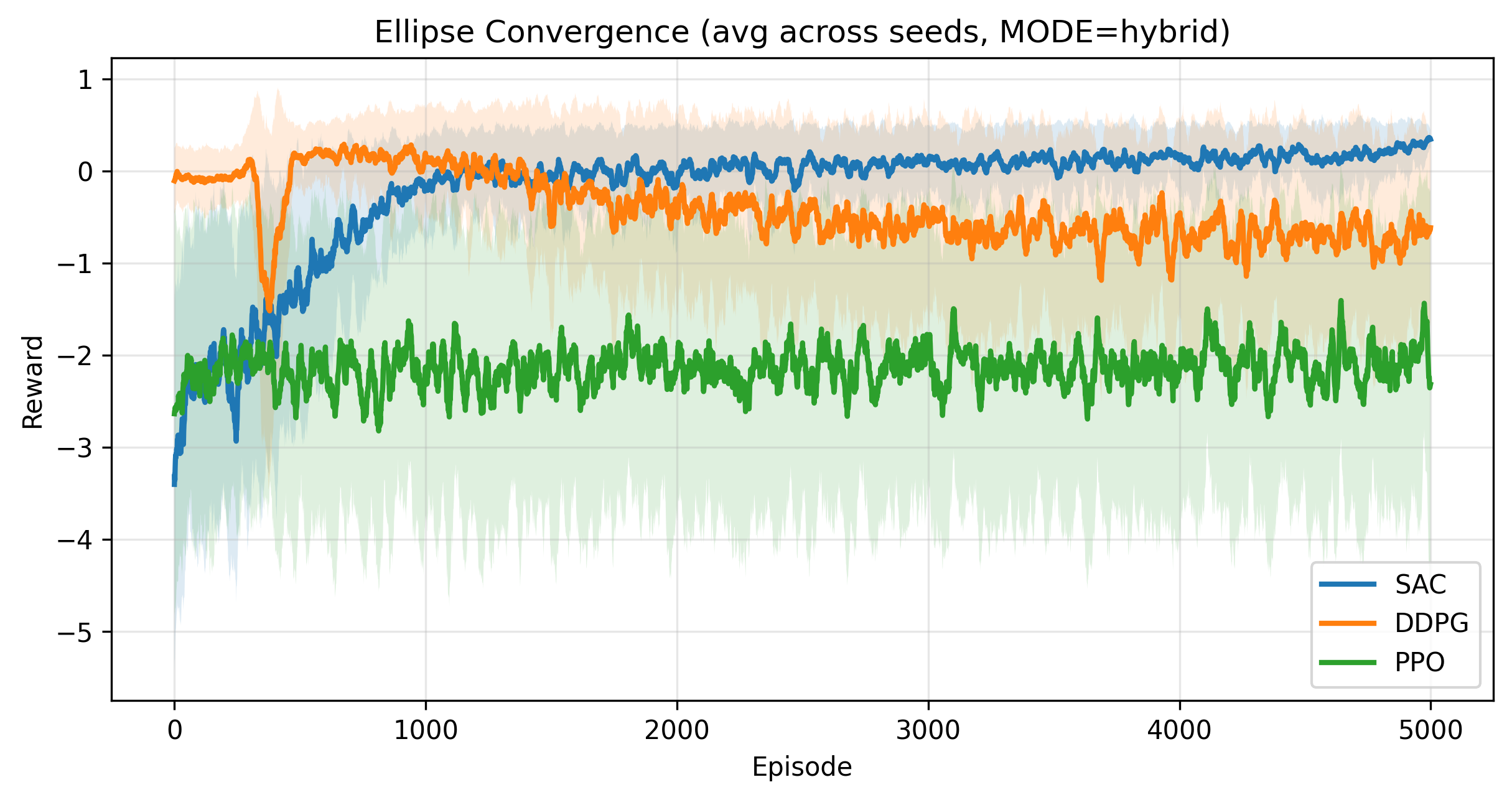}
	\captionof{figure}{Ellipse: training curves averaged across seeds.}
	\label{fig:ell-conv}
\end{center}

\vspace{8pt}

\begin{center}
	\includegraphics[width=0.92\columnwidth,height=0.28\textheight,keepaspectratio]{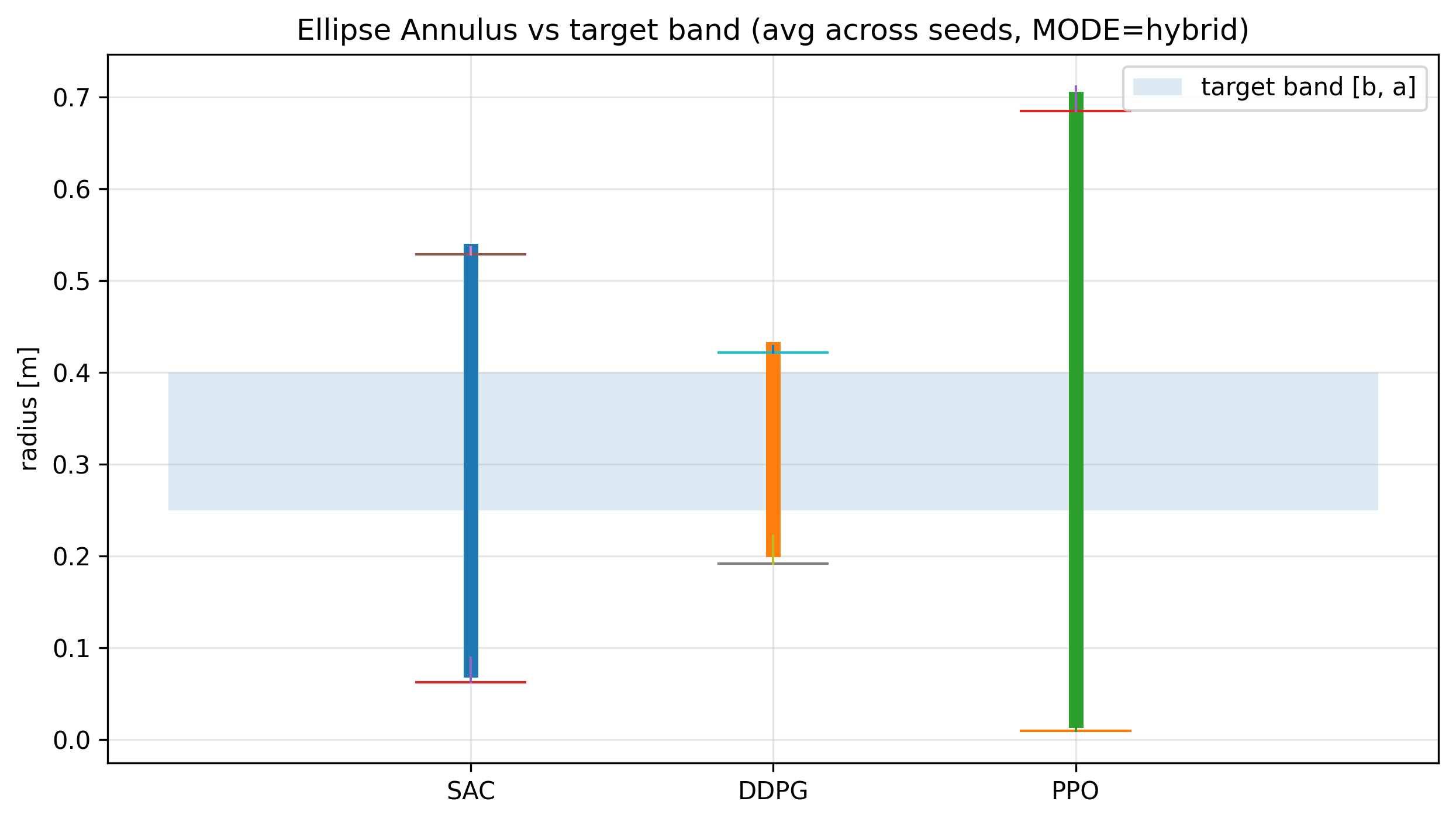}
	\captionof{figure}{Ellipse (hybrid reward): learned reachable annulus $[r_{\min},r_{\max}]$ averaged across seeds for SAC, DDPG, and PPO, compared against the target task band $[b,a]$ (shaded). Vertical bars show the learned inner/outer radii; small horizontal ticks indicate the across-seed variation.}
	\label{fig:ell-annulus}
\end{center}

\begin{center}
	\includegraphics[width=0.92\columnwidth,height=0.28\textheight,keepaspectratio]{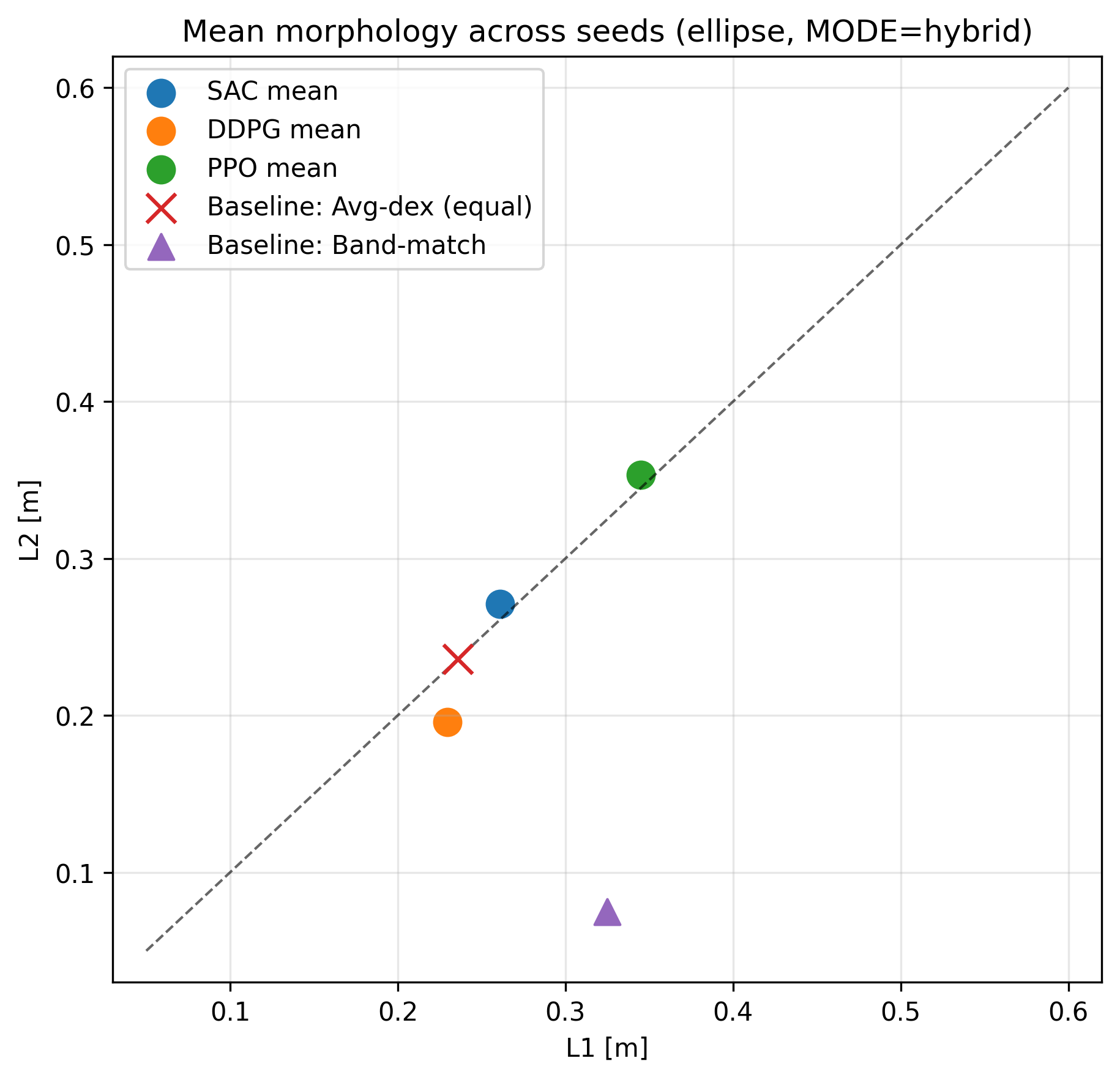}
	\captionof{figure}{Ellipse (hybrid reward): mean morphology across seeds.
		Dashed: $L_1{=}L_2$.}
	\label{fig:ell-mean}
\end{center}


\vspace{6pt}

Figure~\ref{fig:ell-annulus} complements the mean-length plot in Fig.~\ref{fig:ell-mean} by showing how closely each algorithm matches the target band $[b,a]$. 
DDPG produces the tightest annulus, with $r_{\min}$ and $r_{\max}$ aligned near the band edges. 
SAC tends to overshoot the outer radius and slightly undershoot the inner radius, yielding a wider annulus than required but with stable behavior across seeds. 
PPO exhibits the largest spread, with a very small $r_{\min}$ and a large $r_{\max}$, indicating a conservative morphology that spans well beyond the desired band under the same 5{,}000-episode budget. 

These annulus patterns are consistent with the training curves in Fig.~\ref{fig:ell-conv}: DDPG reaches higher rewards early but shows large fluctuations, which explains the larger variance across seeds in the final geometries. 
SAC learns more slowly but converges smoothly, leading to more consistent morphologies even if the solution is slightly less band-optimal. 
PPO shows the slowest and lowest reward convergence, which correlates with its broader and less task-focused annulus in Fig.~\ref{fig:ell-annulus}. 
Together, the learning curves and the final annuli confirm that DDPG is more exploitative but less stable, whereas SAC offers a steadier trade-off between reward and geometric feasibility.

\vspace{6pt}

\pgfplotstableread[col sep=comma,trim cells=true]{combined_ellipse_hybrid_algos_L1L2R.csv}\ellsum

\pgfplotstablegetelem{0}{L1_mean}\of{\ellsum}\edef\ELLSACLoneMean{\pgfplotsretval}
\pgfplotstablegetelem{0}{L1_std}\of{\ellsum}\edef\ELLSACLoneStd{\pgfplotsretval}
\pgfplotstablegetelem{0}{L2_mean}\of{\ellsum}\edef\ELLSACLtwoMean{\pgfplotsretval}
\pgfplotstablegetelem{0}{L2_std}\of{\ellsum}\edef\ELLSACLtwoStd{\pgfplotsretval}
\pgfplotstablegetelem{0}{R_mean}\of{\ellsum}\edef\ELLSACRMean{\pgfplotsretval}
\pgfplotstablegetelem{0}{R_std}\of{\ellsum}\edef\ELLSACRStd{\pgfplotsretval}
\pgfplotstablegetelem{1}{L1_mean}\of{\ellsum}\edef\ELLDDPGLoneMean{\pgfplotsretval}
\pgfplotstablegetelem{1}{L1_std}\of{\ellsum}\edef\ELLDDPGLoneStd{\pgfplotsretval}
\pgfplotstablegetelem{1}{L2_mean}\of{\ellsum}\edef\ELLDDPGLtwoMean{\pgfplotsretval}
\pgfplotstablegetelem{1}{L2_std}\of{\ellsum}\edef\ELLDDPGLtwoStd{\pgfplotsretval}
\pgfplotstablegetelem{1}{R_mean}\of{\ellsum}\edef\ELLDDPGRMean{\pgfplotsretval}
\pgfplotstablegetelem{1}{R_std}\of{\ellsum}\edef\ELLDDPGRStd{\pgfplotsretval}
\pgfplotstablegetelem{2}{L1_mean}\of{\ellsum}\edef\ELLPPOLOneMean{\pgfplotsretval}
\pgfplotstablegetelem{2}{L1_std}\of{\ellsum}\edef\ELLPPOLOneStd{\pgfplotsretval}
\pgfplotstablegetelem{2}{L2_mean}\of{\ellsum}\edef\ELLPPOLOTwoMean{\pgfplotsretval}
\pgfplotstablegetelem{2}{L2_std}\of{\ellsum}\edef\ELLPPOLOTwoStd{\pgfplotsretval}
\pgfplotstablegetelem{2}{R_mean}\of{\ellsum}\edef\ELLPPOREMean{\pgfplotsretval}
\pgfplotstablegetelem{2}{R_std}\of{\ellsum}\edef\ELLPPOREStd{\pgfplotsretval}

\begin{center}
	\footnotesize
	\captionof{table}{Ellipse (hybrid): per-algorithm summary (mean $\pm$ std across seeds).}
	\label{tab:ellipse-hybrid-wide}
	\begin{tabularx}{\columnwidth}{l *{3}{>{\centering\arraybackslash}X}}
		\toprule
		\textbf{Metric} & \textbf{SAC} & \textbf{DDPG} & \textbf{PPO} \\
		\midrule
		$L_1$ [m] &
		\printnum{\ELLSACLoneMean}$\pm$\printnum{\ELLSACLoneStd} &
		\printnum{\ELLDDPGLoneMean}$\pm$\printnum{\ELLDDPGLoneStd} &
		\printnum{\ELLPPOLOneMean}$\pm$\printnum{\ELLPPOLOneStd} \\
		$L_2$ [m] &
		\printnum{\ELLSACLtwoMean}$\pm$\printnum{\ELLSACLtwoStd} &
		\printnum{\ELLDDPGLtwoMean}$\pm$\printnum{\ELLDDPGLtwoStd} &
		\printnum{\ELLPPOLOTwoMean}$\pm$\printnum{\ELLPPOLOTwoStd} \\
		Reward &
		\printnum{\ELLSACRMean}$\pm$\printnum{\ELLSACRStd} &
		\printnum{\ELLDDPGRMean}$\pm$\printnum{\ELLDDPGRStd} &
		\printnum{\ELLPPOREMean}$\pm$\printnum{\ELLPPOREStd} \\
		\bottomrule
	\end{tabularx}
\end{center}



\endgroup

\subsubsection{Rectangle: results}
\label{sec:rect-results}


\begingroup
\setlength{\abovecaptionskip}{2pt}
\setlength{\belowcaptionskip}{2pt}
\renewcommand{\arraystretch}{0.96}
\setlength{\tabcolsep}{3pt}

\begin{center}
	\includegraphics[width=\columnwidth,height=0.28\textheight,keepaspectratio]{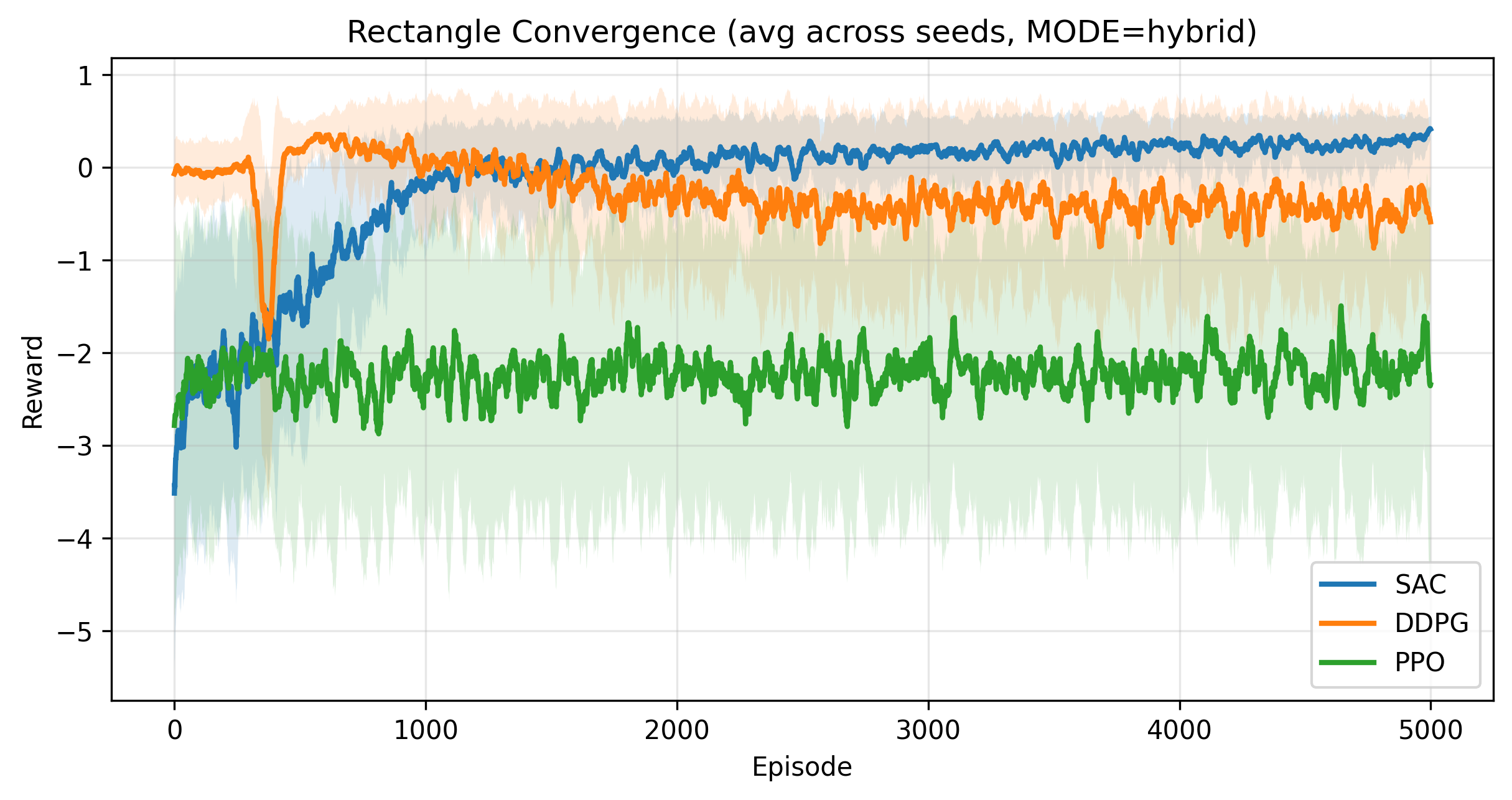}
	\captionof{figure}{Rectangle: training curves averaged across seeds.}
	\label{fig:rect-conv}
\end{center}

\vspace{4pt}

\begin{center}
	\includegraphics[width=0.92\columnwidth,height=0.28\textheight,keepaspectratio]{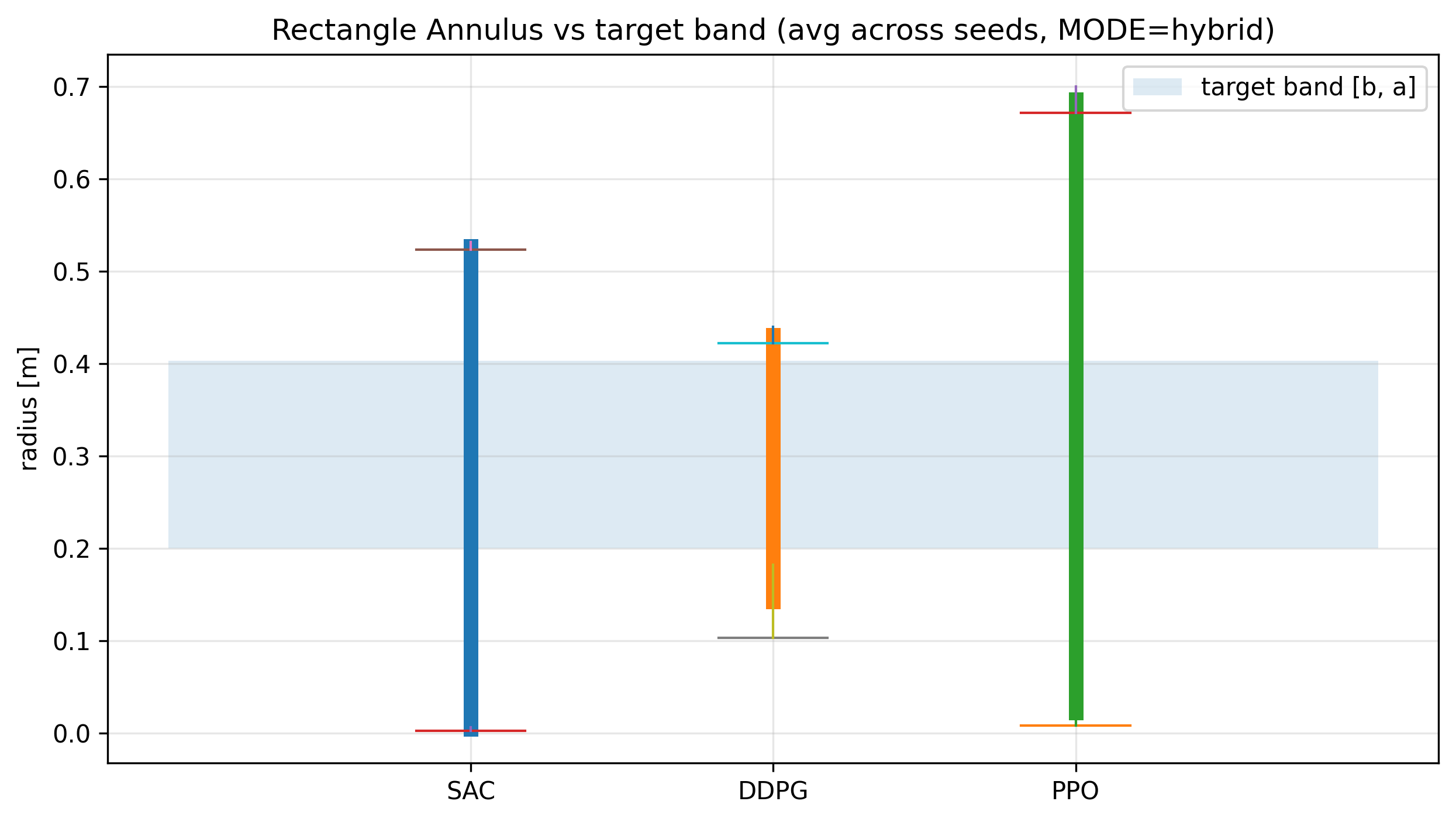}
	\captionof{figure}{Rectangle (hybrid reward): learned reachable annulus $[r_{\min}, r_{\max}]$ averaged across seeds for SAC, DDPG, and PPO, compared against the target task band $[b,a]$ (shaded). Vertical bars show learned inner/outer radii; horizontal ticks denote across-seed variation.}
	\label{fig:rect-annulus}
\end{center}
\begin{center}
	\includegraphics[width=0.92\columnwidth,height=0.28\textheight,keepaspectratio]{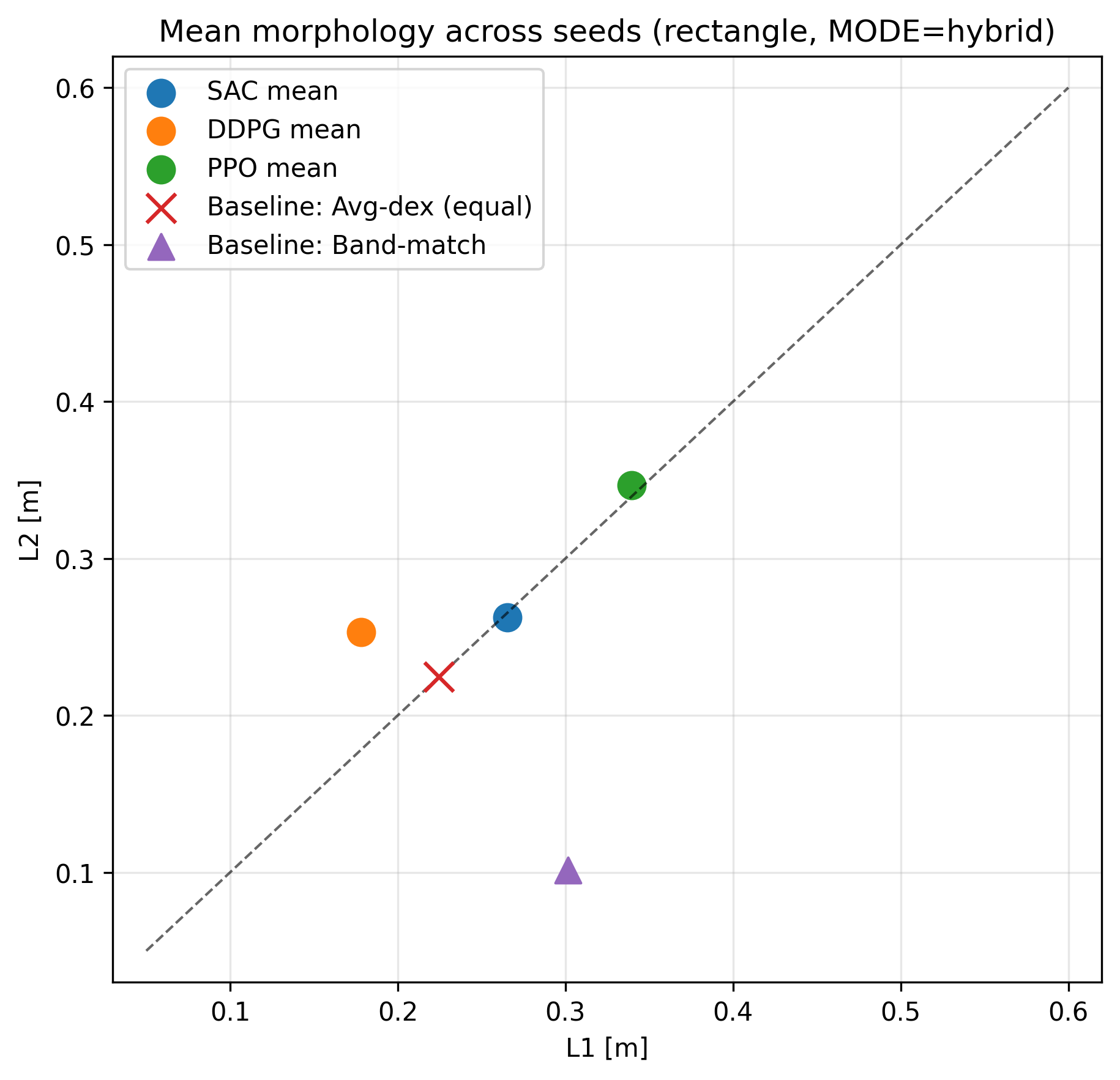}
	\captionof{figure}{Rectangle (hybrid reward): mean morphology across seeds.
		Dashed: $L_1{=}L_2$.}
	\label{fig:rect-mean}
\end{center}

\vspace{4pt}


\vspace{6pt}

Figure~\ref{fig:rect-annulus} shows the same analysis for the rectangular task. 
As in the ellipse case, DDPG produces the narrowest annulus and stays closest to the desired band, although with noticeable variation across seeds. 
SAC again overshoots the outer radius and undershoots the inner radius, resulting in a wider annulus that still overlaps the target region. 
PPO yields the largest spread and tends to cover a much larger radial range than required, reflecting its lower sample efficiency under the same 5{,}000-episode budget.

These annulus outcomes align with the training curves in Fig.~\ref{fig:rect-conv}: DDPG reaches high reward early but with strong oscillations, which explains why its final geometries vary more from seed to seed. 
SAC improves more steadily and converges smoothly, leading to more consistent morphologies even if the final annulus is not the tightest. 
PPO again shows the slowest reward growth and lowest final return, which matches its tendency to produce conservative, over-extended morphologies in Fig.~\ref{fig:rect-annulus}. 
Overall, the rectangular results reinforce the same pattern observed in the ellipse task: DDPG is fast but unstable, SAC is slower but reliable, and PPO is the least sample-efficient in this setting.

\vspace{6pt}
\pgfplotstableread[col sep=comma,trim cells=true]{\detokenize{combined_rectangle_hybrid_algos_L1L2R.csv}}\rectsum

\pgfplotstablegetelem{0}{L1_mean}\of{\rectsum}\edef\RECSACLoneMean{\pgfplotsretval}
\pgfplotstablegetelem{0}{L1_std}\of{\rectsum}\edef\RECSACLoneStd{\pgfplotsretval}
\pgfplotstablegetelem{0}{L2_mean}\of{\rectsum}\edef\RECSACLtwoMean{\pgfplotsretval}
\pgfplotstablegetelem{0}{L2_std}\of{\rectsum}\edef\RECSACLtwoStd{\pgfplotsretval}
\pgfplotstablegetelem{0}{R_mean}\of{\rectsum}\edef\RECSACRMean{\pgfplotsretval}
\pgfplotstablegetelem{0}{R_std}\of{\rectsum}\edef\RECSACRStd{\pgfplotsretval}
\pgfplotstablegetelem{1}{L1_mean}\of{\rectsum}\edef\RECDDPGLoneMean{\pgfplotsretval}
\pgfplotstablegetelem{1}{L1_std}\of{\rectsum}\edef\RECDDPGLoneStd{\pgfplotsretval}
\pgfplotstablegetelem{1}{L2_mean}\of{\rectsum}\edef\RECDDPGLtwoMean{\pgfplotsretval}
\pgfplotstablegetelem{1}{L2_std}\of{\rectsum}\edef\RECDDPGLtwoStd{\pgfplotsretval}
\pgfplotstablegetelem{1}{R_mean}\of{\rectsum}\edef\RECDDPGRMean{\pgfplotsretval}
\pgfplotstablegetelem{1}{R_std}\of{\rectsum}\edef\RECDDPGRStd{\pgfplotsretval}
\pgfplotstablegetelem{2}{L1_mean}\of{\rectsum}\edef\RECPPOLOneMean{\pgfplotsretval}
\pgfplotstablegetelem{2}{L1_std}\of{\rectsum}\edef\RECPPOLOneStd{\pgfplotsretval}
\pgfplotstablegetelem{2}{L2_mean}\of{\rectsum}\edef\RECPPOLOTwoMean{\pgfplotsretval}
\pgfplotstablegetelem{2}{L2_std}\of{\rectsum}\edef\RECPPOLOTwoStd{\pgfplotsretval}
\pgfplotstablegetelem{2}{R_mean}\of{\rectsum}\edef\RECPPOREMean{\pgfplotsretval}
\pgfplotstablegetelem{2}{R_std}\of{\rectsum}\edef\RECPPOREStd{\pgfplotsretval}

\begin{center}
	\footnotesize
	\captionof{table}{Rectangle (hybrid): per-algorithm summary (mean $\pm$ std across seeds).}
	\label{tab:rect-hybrid-wide}
	\begin{tabularx}{\columnwidth}{l *{3}{>{\centering\arraybackslash}X}}
		\toprule
		\textbf{Metric} & \textbf{SAC} & \textbf{DDPG} & \textbf{PPO} \\
		\midrule
		$L_1$ [m] &
		\printnum{\RECSACLoneMean}$\pm$\printnum{\RECSACLoneStd} &
		\printnum{\RECDDPGLoneMean}$\pm$\printnum{\RECDDPGLoneStd} &
		\printnum{\RECPPOLOneMean}$\pm$\printnum{\RECPPOLOneStd} \\
		$L_2$ [m] &
		\printnum{\RECSACLtwoMean}$\pm$\printnum{\RECSACLtwoStd} &
		\printnum{\RECDDPGLtwoMean}$\pm$\printnum{\RECDDPGLtwoStd} &
		\printnum{\RECPPOLOTwoMean}$\pm$\printnum{\RECPPOLOTwoStd} \\
		Reward &
		\printnum{\RECSACRMean}$\pm$\printnum{\RECSACRStd} &
		\printnum{\RECDDPGRMean}$\pm$\printnum{\RECDDPGRStd} &
		\printnum{\RECPPOREMean}$\pm$\printnum{\RECPPOREStd} \\
		\bottomrule
	\end{tabularx}
\end{center}
\vspace*{5pt}


\endgroup

\FloatBarrier

\section{Discussion and future scope}
\label{sec:discussion}

The results above show that a simple contextual-bandit RL loop can recover analytical optima and provide competitive morphologies for non-analytical tasks. 
This section outlines practical scalability considerations, possible extensions and directions for future work that would increase the scope, realism and applicability of the proposed framework.

\begin{itemize}
	\item \textbf{Higher–DoF morphology.} 
	The formulation can be extended from a planar 2R manipulator to an $n$–DoF manipulators design by collecting the design variables into an action vector $a=[L_1,\dots,L_n,\psi]$ and using manipulability $w(\theta)=\sqrt{\det(JJ^\top)}$ with $J\in\mathbb{R}^{m\times n}$. 
	Exhaustive search grows exponentially with $n$, while the RL/bandit approach scales with the action dimension; the dominant per-episode cost remains the IK/Jacobian evaluation over $N$ samples (roughly $O(N\,n^2)$). 
	This cost is amenable to parallelization across samples and seeds, but careful algorithmic choices and dimensionality reduction (e.g., shared encoders or structured parameterizations) are required for large $n$.
	
	\item \textbf{Richer objectives and constraints.} 
	The current work focuses on kinematic dexterity, but actuator limits and dynamics can be incorporated by adding penalties or constraints to the reward. 
	For example, joint-velocity limits follow from $\dot\theta=J^{-1}\dot{x}$ and torque limits from
	$\tau=M(\theta)\ddot\theta+C(\theta,\dot\theta)\dot\theta+g(\theta)$; penalties can aggregate maximum or mean violations along the path. 
	Alternative isotropy measures (e.g., condition number $\kappa(J)$ or task-aligned manipulability) and accuracy/stiffness objectives can be included without changing the learning loop, yielding more physically realistic designs.
	
	\item \textbf{From paths to regions and 3D tasks.} 
	The radial ``band'' idea generalizes naturally by sampling targets from 2-D or 3-D workspace distributions rather than a single path. 
	Orientation tracking can be handled by the $6\times n$ spatial Jacobian, with reward terms that weight translation and rotation according to task priorities. 
	This extension enables design for volumetric workspaces and full SE(3) tasks.
	
	\item \textbf{Sample efficiency and generalization.} 
	Improving data efficiency is crucial for higher complexity. 
	Promising approaches include curricula (loose $\rightarrow$ tight bands), task randomization over shape/size, shared encoders for task context, offline pretraining on synthetic task distributions, and differentiable-kinematics fine-tuning. 
	Such measures improve transfer across tasks and reduce the number of expensive online evaluations.
	
	\item \textbf{Deployment and robustness.} 
	Real-world deployment requires modeling uncertainty in $L_i$, actuator backlash, sensing noise, and fabrication tolerances. 
	Robust objectives — for example optimizing expected performance or tail risk (CVaR) — and validation via a URDF pipeline or hardware-in-the-loop tests are natural next steps to bridge sim-to-real gaps.
	
	\item \textbf{Multi-objective design.} 
	Practical design problems often trade off dexterity, coverage, energy consumption and mass. 
	Instead of producing a single optimum, the framework can be extended to return Pareto fronts (via scalarization, multi-objective RL, or Pareto RL) so practitioners obtain families of morphology candidates that expose trade-offs for downstream selection.
\end{itemize}

Taken together, these directions sketch a roadmap from the present kinematic study towards a comprehensive, dynamic, and robust morphology co-design pipeline suitable for more complex robots and real-world validation.

\section{Conclusion}
\label{sec:conclusion}

We studied morphology optimization for planar manipulators using an analytical,
heuristic, and reinforcement-learning (RL) toolkit. For the centered circle, the
closed-form optimum ($L_1{=}L_2{=}R/\sqrt{2}$, $\theta_2{=}90^\circ$) is recovered
numerically by RL and heuristics, matching the analytical reward
$w_{\mathrm{norm}}(\phi){=}\sin(2\phi)$. For ellipse and rectangle, we introduced a
hybrid reward function that balances normalized dexterity with geometric feasibility via a
band penalty. Across all the algorithms (SAC, DDPG, and PPO), the agent attains competitive or superior
performance to classic search baselines under identical evaluation budgets, with SAC
typically offering the most robust greedy solutions and DDPG learning fastest but with
higher variance.

The main takeaway is that a single, simple contextual-bandit RL loop can handle
non-analytical tasks and constraint trade-offs that are cumbersome for grid search or
black-box heuristics, while remaining consistent with analytical ground truth when it
exists. The scope of the work is limited to  planar kinematics considering absence of
explicit torque/velocity, collision, and orientation objectives. Future work can
scale to higher-DoF and 3-D tasks, incorporate dynamic constraints and multi-objective
criteria, and validate on hardware with robust/sim-to-real training.

\bibliography{ifacconf}             

\nocite{*}

\appendix

\section{Derivations of \(\mathbb{E}[r^2]\)}
\label{app:Er2}

The following expressions for the mean squared radius \(\mathbb{E}[r^2]\) of
the ellipse and rectangle are obtained from standard analytic-geometry
integrations of \(r^2 = x^2 + y^2\) along their respective perimeters
(Weisstein (n.d.)).

\paragraph*{Ellipse.}
With \(x = a\cos t\) and \(y = b\sin t\), the radial distance is
\(r^2 = a^2\cos^2 t + b^2\sin^2 t\).
Uniformly averaging over \(t \in [0, 2\pi]\) gives
\[
\mathbb{E}[r^2]_{\mathrm{ellipse}}
= \frac{1}{2\pi}\!\int_0^{2\pi}\!(a^2\cos^2 t + b^2\sin^2 t)\,dt
= \tfrac{1}{2}(a^2 + b^2).
\]

\paragraph*{Rectangle.}
For an axis-aligned rectangle of width \(w\) and height \(h\),
centered at the origin, the perimeter-weighted average
\(\mathbb{E}[r^2] = \frac{1}{2(w + h)}\int_{\partial\mathcal{R}}(x^2 + y^2)\,ds\)
evaluates to
\[
\mathbb{E}[r^2]_{\mathrm{rect}}
= \frac{w^3 + 3wh^2 + 3hw^2 + h^3}{12(w + h)}.
\]
These results are consistent with the geometric formulations
summarized in (Weisstein (n.d.)).

\section{Heuristic Optimizer Settings}\label{app:heur}
\begin{table}[H]
	\centering
	\caption{Heuristic optimizer settings for 1-D $\phi$ (maximize $f(\phi)=\sin(2\phi)$).}
	\label{tab:heuristic-params}
	\footnotesize
	\setlength{\tabcolsep}{2pt}
	\renewcommand{\arraystretch}{1.06}
	\begin{tabular}{
			@{} >{\raggedright\arraybackslash}p{.26\linewidth}
			@{\hspace{1pt}} >{\raggedright\arraybackslash}p{.25\linewidth}
			@{\hspace{1pt}} >{\raggedright\arraybackslash}p{.25\linewidth}
			@{\hspace{1pt}} >{\raggedright\arraybackslash}p{.25\linewidth} @{}
		}
		\toprule
		& \makecell[l]{\textbf{PSO}} &
		\makecell[l]{\textbf{BO}\\\textbf{(GP+EI)}} &
		\makecell[l]{\textbf{CMA-}\\\textbf{ES}} \\
		\midrule
		Evals (total) & 3630 & 45 & 720 \\
		Iters         & 120  & 40 & 60 \\
		Pop / batch   & $n_p{=}30$ & $n_{\text{init}}{=}5$ & $\lambda{=}12,\ \mu{=}6$ \\
		Seed          & 0 & 0 & 0 \\
		Init sampling & $U[\phi_{\min},\phi_{\max}]$ & $U[\phi_{\min},\phi_{\max}]$ & $m_0=\tfrac{\phi_{\min}+\phi_{\max}}{2}$ \\
		Init scale    & $v\!\sim\!U[-0.05,0.05]$ & --- & $\sigma_0=0.20(\phi_{\max}-\phi_{\min})$ \\
		Update core   & $v\!\leftarrow\!\omega v+c_1 r_1(p{-}x)+c_2 r_2(g{-}x)$
		& GP (RBF $\ell{=}0.12$, noise $10^{-10}$) + EI ($\xi{=}0.01$), grid 2000
		& rank-$\mu$; $\sigma$ clip $[10^{-4},\,0.5(\phi_{\max}-\phi_{\min})]$ \\
		Params        & \makecell[l]{$\omega{=}0.72$\\$c_1{=}c_2{=}1.49$}
		& lengthscale $0.12$ & elitist \\
		Bounds        & clip to $[\phi_{\min},\phi_{\max}]$
		& grid inside bounds & clip to bounds \\
		\bottomrule
	\end{tabular}
\end{table}

\FloatBarrier    

\end{document}